\documentclass[final]{cvpr}

\usepackage{times}
\usepackage{epsfig}
\usepackage{graphicx}
\usepackage{amsmath}
\usepackage{amssymb}
\usepackage{multirow}
\usepackage{algorithm} 
\usepackage{algpseudocode}

\usepackage[pagebackref=true,breaklinks=true,colorlinks,bookmarks=false]{hyperref}

\def\cvprPaperID{937} 
\def\confYear{CVPR 2021}

\begin{document}

\title{Hardness Sampling for Self-Training Based Transductive Zero-Shot Learning}

\author{Liu Bo,\textsuperscript{\rm 1}\textsuperscript{\rm 2}
Qiulei Dong,\textsuperscript{\rm 1}\textsuperscript{\rm 3}\textsuperscript{\rm 4}
Zhanyi Hu\textsuperscript{\rm 1}\textsuperscript{\rm 2}\\
\textsuperscript{\rm 1}National Laboratory of Pattern Recognition, CASIA\\
\textsuperscript{\rm 2}School of Future Technology, UCAS \
\textsuperscript{\rm 3}School of Artificial Intelligence, UCAS\\
\textsuperscript{\rm 4}Center for Excellence in Brain Science and Intelligence Technology, CAS\\
{\tt\small liubo2017@ia.ac.cn \{qldong,huzy\}@nlpr.ia.ac.cn,Corresponding author:Qiulei Dong}
}

\maketitle

\begin{abstract}
  Transductive zero-shot learning (T-ZSL) which could alleviate the domain shift problem in existing ZSL works, has received much attention recently. However, an open problem in T-ZSL: how to effectively make use of unseen-class samples for training, still remains. Addressing this problem, we first empirically analyze the roles of unseen-class samples with different degrees of hardness in the training process based on the uneven prediction phenomenon found in many ZSL methods, resulting in three observations. Then, we propose two hardness sampling approaches for selecting a subset of diverse and hard samples from a given unseen-class dataset according to these observations. The first one identifies the samples based on the class-level frequency of the model predictions while the second enhances the former by normalizing the class frequency via an approximate class prior estimated by an explored prior estimation algorithm. Finally, we design a new Self-Training framework with Hardness Sampling for T-ZSL, called STHS, where an arbitrary inductive ZSL method could be seamlessly embedded and it is iteratively trained with unseen-class samples selected by the hardness sampling approach. We introduce two typical ZSL methods into the STHS framework and extensive experiments demonstrate that the derived T-ZSL methods outperform many state-of-the-art methods on three public benchmarks. Besides, we note that the unseen-class dataset is separately used for training in some existing transductive generalized ZSL (T-GZSL) methods, which is not strict for a GZSL task. Hence, we suggest a more strict T-GZSL data setting and establish a competitive baseline on this setting by introducing the proposed STHS framework to T-GZSL.
\end{abstract}

\section{Introduction}
Recently, zero-shot learning (ZSL), which aims to recognize unseen-class samples given a set of labeled seen-class samples and the semantic features of both the seen and unseen classes, has attracted increasing attention in the fields of machine learning and computer vision. The key to ZSL is to learn an appropriate mapping between visual and semantic features, which could be adapted to the unseen-class domain.

According to whether the unlabeled unseen-class samples are available for training, the existing ZSL methods could be divided into two categories: inductive ZSL (I-ZSL) methods~\cite{Frome13DeViSE,zhang2017DEM,Xian18FCLSWGAN,li2019LiGAN,zhu2019ABP,jiang2019TCN,keshari2020OCDZSL,Vyas2020LrGAN,liu2020APNet, Ma2020DE-VAE, huynh2020DAZLE} where only the labeled seen-class samples are available for training and transductive ZSL (T-ZSL) methods~\cite{verma2017GFZSL,Song18QFSL,xian2019f-VAEGAN,li2019GXE,paul2019SABR,YeG19PREN,wan2019vsc,wu2020SDGN,Narayan2020TF-VAEGAN} where both the labeled seen-class samples and the unlabeled unseen-class samples are available for training. As pointed out in~\cite{Kodirov15uda,wan2019vsc,YeG19PREN,Narayan2020TF-VAEGAN} that the I-ZSL methods generally suffer from the domain shift problem that the learned model from the seen-class domain may not be suitable for the unseen-class one. In contrast to the I-ZSL methods, the T-ZSL methods could alleviate the domain shift problem to some extent due to the used unseen-class samples. However, it still remains the following open problem for T-ZSL: how to make use of unlabeled unseen-class samples for training more effectively?

Addressing this problem, we firstly investigate the prediction accuracies of unseen classes on the public dataset by several popular I-ZSL methods, finding an uneven prediction phenomenon that for each of these methods, its prediction accuracies of some unseen classes are considerably different from those of the other unseen classes, that is to say, it is relatively harder to classify some unseen classes but easier to classify the others. The uneven prediction phenomenon encourages us to further investigate the roles of unseen-class samples with different degrees of classification hardness in the training process, resulting in three observations: When unseen-class samples are available for training, (1) samples from `hard' classes are more effective for improving the performance of a given ZSL method than those from `easy' classes; (2) the predicted pseudo labels of the hard classes are less noisy than those of the easy classes; (3) the diversity of hard classes is beneficial to improve the performance of a given ZSL method.

According to the above observations, we explore two hardness sampling approaches for automatically selecting a subset of diverse and hard samples from the given set of unseen-class samples. The first one selects unseen-class samples according to the class-level frequency of the model predictions, while the second uses an explored prior estimation algorithm to estimate the approximate prior of unseen classes, and then selects unseen-class samples according to the normalized class-level frequency by the approximate prior. Finally, we design a new Self-Training framework with Hardness Sampling for T-ZSL, called STHS, where an arbitrary I-ZSL method could be seamlessly embedded and it is iteratively updated with an automatically selected unseen-class subset by one of the two explored hardness sampling approaches.

A popular data setting in the transductive generalized ZSL(T-GZSL) used in many existing works~\cite{paul2019SABR,xian2019f-VAEGAN,wu2020SDGN} is that the testing unseen-class dataset is separately used for training, implying that it has been known whether a given sample is an unseen-class one or not. However, evaluations on this setting are not able to strictly reflect the ability of a T-GZSL method for handling the GZSL task, because in a GZSL task we only know that the testing dataset contains both seen-class and unseen-class data, but we indeed do not know whether a given sample belongs to the unseen classes or not. Hence, we present a more strict T-GZSL data setting where the unseen-class and seen-class samples are indistinguishable in both training and testing stages and establish a competitive baseline on this setting by introducing the proposed STHS framework to T-GZSL.

In sum, our contributions are three-fold:
\begin{itemize}
	\item We empirically find three factors which are helpful for improving the performance of a given ZSL method when unseen-class samples are available for training. According to these factors, two hardness sampling approaches are proposed, which could effectively select diverse and hard samples from the given set of unseen-class samples.
	\item We design the STHS framework for T-ZSL, where various T-ZSL methods could be easily derived. Our experimental results demonstrate that when two typical I-ZSL methods are introduced into the proposed STHS framework respectively, the derived T-ZSL methods achieve superior performances than many state-of-the-art methods.
	\item We propose a more strict data setting to evaluate T-GZSL methods, and establish a competitive baseline on this setting by generalizing the STHS framework to T-GZSL.
\end{itemize}

\section{Related Work}

\subsection{Inductive Zero-Shot Learning}
Existing inductive ZSL methods focused on learning a mapping between visual features and semantic features using seen-class data, which could be roughly divided into two categories: the embedding-based methods and the generative methods. Among the embedding-based methods, some works~\cite{Frome13DeViSE,Akata16ALE,akata2015sje,romera2015ESZSL,xian2016LATEM,socher2013CMT,Xie_2019_AREN,Zhu19SGML,Yu18SGA,liu2019LFGAA} projected visual features into a semantic feature space by a linear or nonlinear function for object classification, which are impeded by the hubness problem. While some other works~\cite{Changpinyo17EXEM,zhang2017DEM,dvbe2020,EPGN2020} proposed to map semantic features into a visual feature space to alleviate the hubness problem. The generative methods~ \cite{Xian18FCLSWGAN,NiZ019dascn,li2019LiGAN,paul2019SABR,YuL19SGL,zhu2019ABP,ding2019MLSE,huang2019GDAN,schonfeld2019CADA-VAE,Zhu18GAZSL,verma2020ZSML,LiuDH20} utilized a conditional GAN to generate fake unseen-class samples and then these fake samples were used to train a classifier to classify real unseen-class samples.

\subsection{Transductive Zero-Shot Learning}
Recently, transductive ZSL methods~\cite{wan2019vsc} have been proposed to alleviate the domain shift problem in the inductive ZSL methods. Fu et al.~\cite{Fu15TMV} proposed a method where visual features and semantic features were projected into a multi-view embedding space and then a hyper-graph was constructed using the unlabeled data for label propagation. Kodirov et al.~\cite{Kodirov15uda} utilized an unsupervised domain adaptation method with sparse coding to alleviate the domain shift problem. Song et al.~\cite{Song18QFSL} proposed a method where the sum of probabilities on unseen classes is encouraged to be relatively larger for unseen-class data. More recently, several generative methods~\cite{Narayan2020TF-VAEGAN,paul2019SABR,xian2019f-VAEGAN,wu2020SDGN} have been proposed, where a conditional generator was used to generate seen/unseen class samples and two discriminators were used to discriminate fake/real seen-class and unseen-class samples respectively. In addition, a few works~\cite{YeG19PREN, li2019GXE} learned a more generalized visual-semantic mapping by iteratively training the model with a fixed number of pseudo-labeled unseen-class samples. It has to be pointed out that although our method also employs an iterative training scheme, our main concern is to investigate the roles of different unseen-class samples in the training process and how to select pseudo-labeled samples more effectively to improve an arbitrary I-ZSL method, which is totally different from~\cite{YeG19PREN, li2019GXE}.

\section{Methodology}
We begin by introducing the definition of ZSL. Assume that we have a seen-class dataset $\mathcal{D}^{S}_{tr} = \{(x_{n}, y_{n})\}_{n=1}^{N}$ for training, where $x_{n} \in \mathbf{R}^{V}$ is a visual feature and $y_{n}$ is the class label of $x_{n}$, belonging to the seen-class label set $Y^{S}$, and $N$ is the sample number. The semantic feature set $E=\{e_{y} \in \mathbf{R}^{S} \mid y \in Y\}$ is also given, where $Y$ is the total class label set which includes not only the seen-class label set $Y^{S}$ but also the unseen-class label set $Y^{U}$ disjoint with $Y^{S}$. Given a test set $X$, in the conventional ZSL, the task is to learn a mapping $F: X \to Y^{U}$ with the training set and semantic feature set, while in the generalized ZSL, the task is to learn a mapping $F: X \to Y$.  We denote the unlabeled unseen-class dataset by $\mathcal{D}^{U}$ and unlabeled seen-class dataset by $\mathcal{D}^{S}$ in the testing stage respectively. For the transductive ZSL, we assume that the testing unseen-class dataset $\mathcal{D}^{U}$ is available in the training stage. For the transductive generalized ZSL, we assume that the testing unseen-class dataset $\mathcal{D}^{U}$ and testing seen-class dataset $\mathcal{D}^{S}$ are both available in the training stage.

In this section, we propose a Self-Training framework with Hardness Sampling (STHS) for T-ZSL, where a base I-ZSL model is iteratively trained with an automatically selected pseudo-labeled unseen-class subset. In the STHS framework, the core problem is how to select pseudo-labeled samples from the given unseen-class dataset for retraining. Addressing this problem, we firstly analyze the uneven prediction phenomenon in $4$ existing ZSL methods and make three observations about the roles of unseen-class samples with different degrees of hardness in the training process. Then, we propose two hardness sampling methods according to these observations. Finally, we introduce the details of the STHS framework.

\begin{figure}[t]
	\centering
	\includegraphics[width=0.9\linewidth]{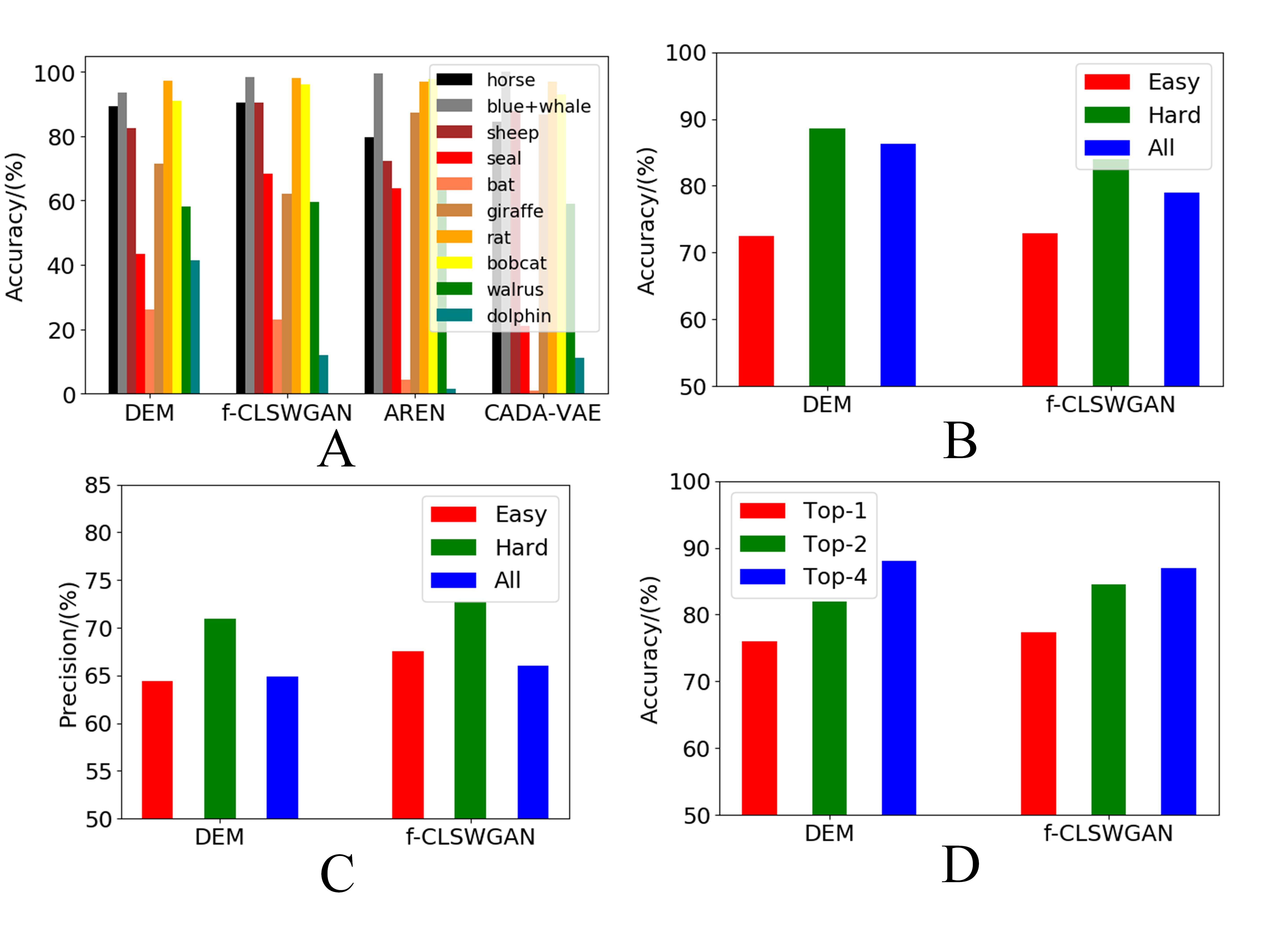}
	\caption{A: the uneven prediction phenomenon. B: the ACCs of models retrained with additional easy-class, hard-class and all-unseen-class data. C: the Precisions of the easy classes, hard classes, and all unseen classes. D: the ACCs of the models retrained with additional Top-$1$, Top-$2$, Top-$4$ hard-class data.}
	\label{fig1}
\end{figure}

\subsection{Influence of Unseen-Class Samples on Training}
Here we empirically investigate the influence of unseen-class samples on training ZSL models when the unlabeled unseen-class dataset is available. At the very beginning, we calculate the per-class accuracies on AWA2~\cite{Xian17Comprehensive} of $4$ typical ZSL methods, including DEM~\cite{zhang2017DEM} (a semantic-visual embedding-based method), AREN~\cite{Xie_2019_AREN} (a visual-semantic embedding-based method), f-CLSWGAN~\cite{Xian18FCLSWGAN} (a generative method), and CADA-VAE~\cite{schonfeld2019CADA-VAE} (a generative method). The results are shown in Figure~\ref{fig1} A. From Figure~\ref{fig1} A, we find an \textbf{uneven prediction phenomenon} that some unseen classes have very high accuracies while the others have very low accuracies. Here, we name the unseen classes with high accuracies as easy classes and those with low accuracies as hard classes. The uneven prediction phenomenon encourages us to further investigate the roles of unseen-class samples with different degrees of hardness in the training process from $3$ aspects as:

\textbf{Are Hard and Easy Classes Equally Important for Training?} To answer this question, we perform a contrastive experiment on AWA2~\cite{Xian17Comprehensive} using DEM~\cite{zhang2017DEM} and f-CLSWGAN~\cite{Xian18FCLSWGAN}. Specifically, for each method, we firstly train a model with the training set (only including seen-class data) and then make predictions on the unlabeled unseen-class dataset with the trained model. Then, we compute the accuracies of all unseen classes. According to these per-class accuracies, we determine the Top-$6$ easy classes and Top-$6$ hard classes. Next, we randomly select a fixed number of samples in the following $3$ manners: 1) from the Top-$6$ easy classes, 2) from the Top-$6$ hard classes, 3) from all the unseen classes, and add them into the training set respectively. Finally we retrain three models with the three datasets respectively. Note that the true labels instead of the predicted pseudo labels of unseen-class samples are used in the model retraining since this experiment is to investigate whether the hard and easy classes are equally important, which needs to exclude other potential factors, such as label noise. After retraining, we evaluate the ZSL performance of each retrained model by average per-class Top-$1$ accuracy (ACC). The results of DEM and f-CLSWGAN are reported in Figure~\ref{fig1} B. As shown in Figure~\ref{fig1} B, the red, green, and blue bars show the ACCs of the models retrained with additional easy-class, hard-class, and all-unseen-class data respectively. We find that the model retrained with additional hard-class data has the best performance for both methods. This demonstrates that \textbf{the hard classes play a more important role for improving the performance of a given ZSL method than the easy classes.}

\textbf{Do Pseudo Labels of Hard and Easy Classes Have Equal Level of Noise?} In the above experiment, we retrain the models using the true labels of unseen-class samples. However, in practice, what we could indeed have are only the predicted pseudo labels. Here we investigate whether the pseudo labels of easy-class and hard-class samples have equal level of noise? To answer this question, we evaluate the noise level of pseudo labels predicted by DEM~\cite{zhang2017DEM} and f-CLSWGAN~\cite{Xian18FCLSWGAN} on AWA2~\cite{Xian17Comprehensive} using \textbf{Precision} metric. In the multi-class case, precision of the class C is equal to the ratio of the true C-class samples to all the samples predicted as the class C. Specifically, for each method, we firstly train a model with the training set and then obtain the predictions on unseen-class dataset using the trained model. Then, we compute the per-class precisions and per-class accuracies, and obtain the Top-$6$ easy classes and Top-$6$ hard classes according to the per-class accuracies. Finally, we compare the average per-class precision (Precision) of the Top-$6$ easy classes, Precision of the Top-$6$ hard classes, and Precision of all unseen classes. The results of DEM and f-CLSWGAN are shown in Figure~\ref{fig1} C, where the red, green, and blue bars show the Precisions of the Top-$6$ easy classes, Top-$6$ hard classes, and all unseen classes respectively. We observe that the Precision of the Top-$6$ hard classes is significantly higher than those of the other two for both methods. This indicates that \textbf{the pseudo labels of hard classes are less noisy than those of the easy classes.}

\textbf{Is Class Diversity Helpful for Training?} We have experimentally demonstrated that the hard-class samples are more important and less noisy than the easy-class samples. The following question is naturally raised: When the number of available unseen-class samples is given, how to select unseen-class samples is more helpful for training, selecting them from a few hardest classes or selecting them from diverse hard classes? Here we analyze the influence of class diversity on improving the ZSL performance by performing an experiment via DEM~\cite{zhang2017DEM} and f-CLSWGAN~\cite{Xian18FCLSWGAN} on AWA2~\cite{Xian17Comprehensive}. Specifically, for each method, we firstly compute the hardness level of each unseen class according to the per-class accuracies. Then, we select three groups of unseen-class samples with the same sample number, which are from the Top-$1$, Top-$2$ and Top-$4$ hard classes respectively. We add the three groups of data into the training set and retrain three models respectively. Finally, we evaluate the ZSL performance of each retrained model by ACC. The results of DEM and f-CLSWGAN are reported in Figure~\ref{fig1} D, where the red, green, blue bars show the ACCs of the model retrained with the additional Top-$1$, Top-$2$ and Top-$4$ hard-class data respectively. We find that the model retrained with the Top-$4$ hard-class data has the best performance for both methods. This demonstrates that \textbf{the diversity of hard classes is beneficial to improve the performance of ZSL methods.}

\subsection{Hardness Sampling}
According to the three observations in Section 3.1, we speculate that \textbf{it is helpful to boost the ZSL performance by selecting pseudo-labeled unseen-class samples from diverse hard classes.} However, the hard classes are identified with true labels of unseen classes in the above experiments, which are unavailable in practice. Hence, the remaining key problem is how to identify the hard-class samples from a given unseen-class dataset. In this section, we explore two hardness sampling approaches to tackle this problem.

\textbf{Class-Frequency Based Selection.} Here we present the class-frequency based selection (CFBS) method to select diverse hard-class samples, consisting of two steps: (1)identifying the hard classes and (2)uniformly sampling the hard-class instances. To identify the hard classes, we propose a CFBS metric. Specifically, suppose we have a model $F$ and its predictions on the unseen-class dataset $\mathcal{D}^{U}$ are denoted by $P$. We firstly compute the class frequency of the predictions $P$ as follows:
\begin{equation}
\label{eq1}
\{f^{1}, \cdots, f^{c}, \cdots, f^{C}\} = \mathcal{H}(P)
\end{equation}
where $\mathcal{H}(\cdot)$ is the histogram function followed by a normalization function, whose bin number is the unseen-class number $C$ and $f^{c}$ is the frequency of pseudo label $c$ predicted by the model $F$. A smaller $f^{c}$ means that the class $c$ is rarely predicted by the model, which to some extent indicates that the class $c$ is not familiar to the model, i.e. it is hard for the model. Hence, we employ the $f^{c}$ as the metric of class hardness and obtain the hardness level of each unseen class as follows:
\begin{equation}
\label{eq2}
\{d^{1}, \cdots, d^{i}, \cdots, d^{C}\} = \mathrm{argsort} (\{f^{1}, \cdots, f^{c}, \cdots, f^{C}\})
\end{equation}
where $d^{i}$ is the $i$-th hardest class, and $\mathrm{argsort}(\cdot)$ ranks the inputs in an ascending order and return the indexes of sorted inputs. According to Equation \ref{eq2}, we select the Top-$K$ classes $\{d^{i}\}_{1}^{K}$ as the hard classes. Then, we randomly sample $n$ instances from each hard class $d^{i}$ with replacement and finally obtain a subset $\mathcal{\bar{D}}^{U}$ of pseudo-labeled unseen-class data which contains $n*K$ instances in total.

Note that the rationality of the CFBS metric stems from the characteristics of the deep neural networks that the networks are prone to predict the inputs as the classes that they are familiar with. In other words, if a class is rarely predicted by a model, it very likely indicates that the model has seen little data or similar data from the class, i.e. this class is a hard class for the model.

\textbf{Prior-Normalized Class-Frequency Based Selection.} The CFBS metric has its shortcoming. In the case of extremely unbalanced dataset, for a class $c$ with very small number of instances, even if the model is very familiar with this class, the metric $f^{c}$ is still very small, i.e. this class will be wrongly taken as a hard class. Similarly, for a class $c$ with very large number of instances, the metric $f^{c}$ is also ineffective. To overcome this shortcoming, we propose the prior-normalized class-frequency based selection (PN-CFBS) method, which enhances the CFBS method by replacing the CFBS metric by a PN-CFBS metric. Specifically, suppose we have the prior probability $p^{c}$ of the class $c$ and its class frequency $f^{c}$ can be obtained using Equation \ref{eq1}, we define the prior-normalized class frequency of the class $c$ as $\hat{f}^{c} = \frac{f^{c}}{p^{c}}$. Then, we obtain the hardness level of each unseen class $\{\hat{d}^{i}\}_{1}^{C}$ in the same way with Equation \ref{eq2}. Similar to the CFBS method, we could identify the hard classes and select the pseudo-labeled unseen-class subset $\mathcal{\bar{D}}^{U}$.

\begin{figure}[t]
	\centering
	\includegraphics[width=0.8\linewidth]{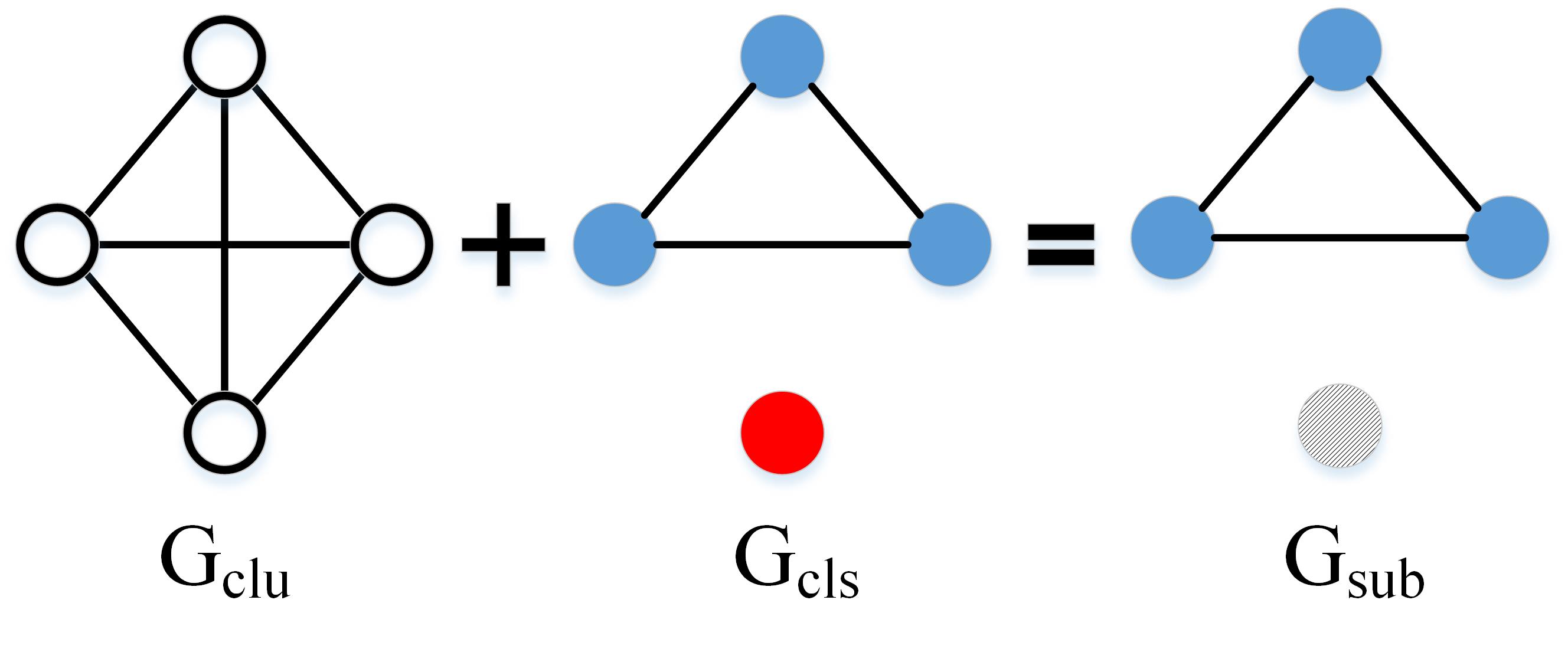}
	\caption{Illustration of the classification-clustering consistency based approximate prior estimation.}
	\label{fig2}
\end{figure}

However, the class prior in the PN-CFBS metric is usually unknown for most datasets, Hence, we propose a classification-clustering consistency based method (3C method) to estimate the approximate class prior for the PN-CFBS method. Specifically, we firstly perform clustering on the unseen-class dataset and obtain the clustering predictions $P_{clu}$. At the same time, we classify the unseen-class dataset with the model $F$ and obtain the classification predictions $P_{cls}$. Then, as shown in Figure~\ref{fig2}, we construct two prediction graphs $G_{clu}$ and $G_{cls}$ using $P_{clu}$ and $P_{cls}$ respectively, where if and only if the nodes have the same label, they will be connected by edges. Next, we construct a sub-graph $G_{sub}$ based on $G_{clu}$ and $G_{cls}$ in the manner that if and only if one edge consistently exists in both $G_{clu}$ and $G_{cls}$, the edge and the corresponding nodes will be preserved in $G_{sub}$. As a result, the sub-graph $G_{sub}$ has two properties: clustering property and semantic property. The clustering property comes from $G_{clu}$, which could reflect the class prior information. As shown in Figure~\ref{fig2}, the sub-graph $G_{sub}$ keeps most nodes in the cluster. The semantic property comes from $G_{cls}$, which gives the nodes in $G_{sub}$ semantic pseudo labels. As shown in Figure~\ref{fig2}, the colors of the nodes in $G_{sub}$ show the pseudo labels. Suppose the pseudo labels of the nodes in $G_{sub}$ are denoted by $P_{sub}$, we finally use $P_{sub}$ to estimate the approximate prior for unseen classes as follow:
\begin{equation}
\label{eq4}
\{\hat{p}^{1}, \cdots, \hat{p}^{c}, \cdots, \hat{p}^{C}\} = \mathcal{H}(P_{sub})
\end{equation}
where $\hat{p}^{c}$ is the estimated approximate prior of the class $c$. In practice, $\hat{p}^{c}$ will be used to compute the PN-CFBS metric via $\hat{f}^{c} = \frac{f^{c}}{\hat{p}^{c}}$. Note that the proposed 3C method does not devote itself to estimate the exact class prior of a dataset, it is rather designed to work together with the PN-CFBS method for identifying hard classes, where the class prior is not necessary to be very exact, since the PN-CFBS method is only required to compute the order of the prior-normalized class frequencies instead of their exact values.

Compared with the CFBS metric, the PN-CFBS metric is effective on the extremely unbalanced dataset. However, we regard the PN-CFBS metric as a supplement for the CFBS metric when faced with extremely unbalanced dataset because: 1)the CFBS metric has considerable ability to deal with relatively unbalanced dataset since the CFBS method does not require the exact values of class frequencies, but their order; 2)the estimation of class prior for a dataset is a tough open problem, which affects the performance of PN-CFBS metric to some extent. 

\subsection{STHS Framework}
In this section, we introduce the overall STHS framework. Suppose we have a base I-ZSL model $F_{0}$ at the beginning and the whole training process includes $T$ iterative steps. At the $t$-th ($t \in \{1, 2, \cdots, T\}$) step, we firstly train the model $F_{t}$ with the training set $\mathcal{D}_{t}$ which consists of two parts: the seen-class dataset $\mathcal{D}^{S}_{tr}$ and the selected pseudo-labeled unseen-class subset $\mathcal{\bar{D}}^{U}_{t}$. Then, we make predictions on the unseen-class dataset $\mathcal{D}^{U}$ using the trained model $F_{t}$ and these predictions $P_{t}$ are used to select a new pseudo-labeled unseen-class subset $\mathcal{\bar{D}}^{U}_{t+1}$ via the proposed hardness sampling approaches to update the training set as $\mathcal{D}_{t+1}$ which is used to re-train the base model at the $t+1$-th step. Considering that the pseudo labels contain more noise at the beginning of iterative training, we select the pseudo-labeled unseen-class subset in a step-wise incremental manner. Specifically, we select $[M/T]*t$ samples at the $t$-th step, where $M$ is the number of samples in $\mathcal{D}^{U}$ and $[\cdot]$ is a down-round function. The overall algorithm is presented at Algorithm \ref{alg1}. Note that an arbitrary I-ZSL method could be embedded into the proposed STHS framework, here we introduce two typical ZSL models, S2V (an embedding-based method) and WGAN (a generative method) as the base models in the STHS framework, and the derived methods are named as STHS-S2V and STHS-WGAN respectively. The overall diagrams of STHS-S2V and STHS-WGAN and more details about S2V and WGAN are presented in the Appendices.
\begin{algorithm}[t]
	\caption{STHS}  
	\label{alg1}
	\begin{algorithmic}[1]  
		\Require  
		$\mathcal{D}^{S}_{tr}$, $\mathcal{D}^{U}$;   
		\Ensure  
		Predictions ${P}_{T}$ on $\mathcal{D}^{U}$; 
		\State Initialization: train the base model $F_{0}$ with $\mathcal{D}^{S}_{tr}$, obtain $\mathcal{D}_{1}$ using CFBS or PN-CFBS;
		\For{t=1 to T}
		\State Re-train $F_{t}$ with the training set $\mathcal{D}_{t}$;
		\State Make predictions $P_{t}$ on $\mathcal{D}^{U}$ using the trained $F_{t}$;
		\State Update the training set as $\mathcal{D}_{t+1}$ using CFBS or PN-CFBS based on $P_{t}$;
		\EndFor \\
		\Return ${P}_{T}$;
	\end{algorithmic}  
\end{algorithm}

\section{Experimental Results}

\subsection{Experimental Setup}
\textbf{Datasets and Evaluation Protocol.} We evaluate the proposed STHS-S2V and STHS-WGAN on three public benchmarks, including AWA2~\cite{Xian17Comprehensive}, CUB~\cite{WahCUB_200_2011}, SUN~\cite{patterson2012sun}. AWA2 contains 37,322 images from 50 animal classes and each class is annotated with 85 attributes. CUB contains 11,788 images from 200 fine-grained birds species and 312 attributes are annotated for each species. SUN includes 14,340 images belonging to 717 scene classes and each class is annotated with 102 attributes. As done in most existing methods, the class-level attributes are used as semantic features in the proposed methods. The visual features which are extracted by the ImageNet1000 pre-trained ResNet101~\cite{He16resnet} are used as model inputs. We evaluate the proposed methods with the the proposed split (PS)~\cite{Xian17Comprehensive}. The standard split (SS) is not used since some unseen classes in SS have been seen by the ImageNet1000 pre-trained ResNet101. In the conventional ZSL, average per-class Top-$1$ accuracy (ACC) on unseen classes is adopted to evaluate performance. In the generalized ZSL, ACCs on unseen classes and seen classes are firstly computed and then the harmonic mean of the two ACCs is computed to evaluate performance.
 
\textbf{Implementation Details and Comparative Methods.} The based models, S2V and WGAN of our methods STHS-S2V and STHS-WGAN are implemented by slightly simplifying DEM~\cite{zhang2017DEM} and f-CLSWGAN~\cite{Xian18FCLSWGAN} respectively, which are detailedly introduced in the Appendices. The iterative step $T$ of STHS-S2V are $\{4, 8, 11\}$ on AWA2, CUB, and SUN respectively, while $T$ of STHS-WGAN are $\{4, 6, 4\}$ on AWA2, CUB, and SUN respectively. The number of hard classes $K$ are the same for both methods, which are $\{8, 45, 54\}$ on AWA2, CUB, and SUN respectively. For the clustering of unseen-class data, we firstly perform dimension reduction on them via t-SNE~\cite{Maaten2008tsne} technique and then perform clustering with Gaussian Mixture Model (GMM). The reduced dimension for AWA2, CUB, and SUN are $\{3, 5, 5\}$ respectively. The components of GMM for AWA2, CUB, and SUN are their number of unseen classes, i.e. $\{10, 50, 72\}$ respectively. At each iterative step, STHS-S2V is trained by $\{5,30,30\}$ epoches with learning rate as $\{0.001,0.0005,0.001\}$ and batch size as $\{128,128,128\}$ on AWA2, CUB, and SUN respectively. STHS-WGAN is trained by $\{20,40,40\}$ epoches on AWA2, CUB, and SUN respectively, with same learning rate $0.0002$ and same batch size $128$. The fake features generated in STHS-WGAN are used to train a feature classifier by $30$ epoches with learning rate as $0.0002$ and batch size as $256$ on all the three datasets.Our code is available at the Github website~\footnote{https://github.com/flywithcloud/STHS}

The proposed STHS-S2V and STHS-WGAN are compared with the state-of-the-art ZSL methods, including $14$ most recent competitive inductive ZSL methods: DEVISE\cite{Frome13DeViSE}, LATEM\cite{xian2016LATEM}, SAE\cite{Kodirov17SAE}, DEM\cite{zhang2017DEM}, f-CSLWGAN\cite{Xian18FCLSWGAN}, LiGAN\cite{li2019LiGAN}, ABP\cite{zhu2019ABP}, TCN\cite{jiang2019TCN}, OCD-GZSL\cite{keshari2020OCDZSL}, APNet\cite{liu2020APNet}, DE-VAE\cite{Ma2020DE-VAE}, LsrGAN\cite{Vyas2020LrGAN}, DVBE\cite{dvbe2020}, DAZLE\cite{huynh2020DAZLE} and $13$ competitive transductive ZSL methods: ALE-tran\cite{Akata16ALE}, GFZSL\cite{verma2017GFZSL}, DSRL\cite{Ye2017DSRL}, QFSL\cite{Song18QFSL}, GMN\cite{sariyildiz2019GMN}, f-VAEGAN-D2\cite{xian2019f-VAEGAN}, GXE\cite{li2019GXE}, SABR-T\cite{paul2019SABR}, PREN\cite{YeG19PREN}, VSC\cite{wan2019vsc}, SDGN\cite{wu2020SDGN}, ADA\cite{Khare2020ADA}, TF-VAEGAN\cite{Narayan2020TF-VAEGAN}.

\subsection{Conventional ZSL Results}
Here we evaluate the proposed STHS-S2V and STHS-WGAN in the conventional ZSL on AWA2, CUB and SUN with PS data split and compare them with $25$ existing methods. The corresponding results are reported in Table~\ref{tab1} where the methods marked by $\mathcal{I}$ are I-ZSL methods while those marked by $\mathcal{T}$ are T-ZSL methods. The results of comparative methods are cited from the original papers or public results~\cite{Xian17Comprehensive}. Since the base I-ZSL methods, S2V and WGAN are obtained by slightly simplifying DEM~\cite{zhang2017DEM} and f-CLSWGAN~\cite{Xian18FCLSWGAN} respectively, we do not report their results due to the limited space. As seen from Table~\ref{tab1}, STHS-WGAN significantly outperforms the comparative methods in most cases and STHS-S2V also achieves comparable performances with the most recent state-of-the-art methods. Specifically, STHS-WGAN achieves improvements about $1.5\%$ and $2.5\%$ on AWA2 and CUB respectively. All the improvements demonstrate that the proposed methods can effectively select pseudo-labeled unseen-class samples for transductive learning. In addition, the performance differences between STHS-S2V and STHS-WGAN tell us that the base model has an effect on the final performance of the STHS framework. We also note that STHS-WGAN achieves a relatively inferior performance on SUN. This is because that the base model in STHS-WGAN has relatively inferior performance on SUN.
\begin{table}[t]
	\centering
	\caption{Comparative results (ACC) in the conventional ZSL setting on AWA2, CUB and SUN.}
	\resizebox{.65\columnwidth}{!}{
	\begin{tabular}{c|c|ccc}
		\hline
		& Method&   AWA2& CUB& SUN \\
		\hline
		\multirow{12}{*}{$\mathcal{I}$}& DEVISE&    59.7& 52.0& 56.5 \\
		& LATEM&     55.8& 49.3& 55.3 \\
		& SAE&       54.1& 33.3& 40.3 \\
		& DEM&       67.1& 51.7& 61.9 \\
		& f-CSLWGAN&   -& 57.3& 60.8 \\
		& LiGAN&     -& 58.8& 61.7 \\
		& ABP&       70.4& 58.5& 61.5 \\
		& TCN&       71.2& 59.5& 61.5 \\
		& OCD-GZSL&  71.3& 60.3& 63.5\\
		& APNet&     68.0& 57.7& 62.3 \\
		& DE-VAE&    69.3& 63.1& 64.0 \\
		& LsrGAN&    -& 60.3& 62.5 \\
		\hline
		\multirow{13}{*}{$\mathcal{T}$}& ALE-tran&    70.7& 54.5& 55.7 \\
		& GFZSL&       78.6& 50.0& 64.0 \\
		& DSRL&        72.8& 48.7& 56.8 \\
		& QFSL&        79.7& 72.1& 58.3 \\
		& GMN&         -& 64.6& 64.3 \\
		& f-VAEGAN-D2&  89.8& 71.1& 70.1 \\
		& GXE&         83.2& 61.3& 63.5 \\
		& SABR-T&      88.9& 74.0& 67.5 \\
		& PREN&        78.6& 66.4& 62.8 \\
		& VSC&         81.7& 71.0& 62.2 \\
		& ADA&         78.6& -& 65.5 \\
		& SDGN&        93.4& 74.9& 68.4 \\
		& TF-VAEGAN&   92.1& 74.7& \textbf{70.9} \\
		\hline
		& STHS-S2V(Ours)&        91.4& 71.2& \textbf{70.9} \\
		& STHS-WGAN(Ours)&        \textbf{94.9}& \textbf{77.4}& 67.5 \\
		\hline
	\end{tabular}
	}
	\label{tab1}
\end{table}

\subsection{Generalized ZSL Results}
We perform the generalized ZSL (GZSL) with the derived methods from the STHS framework on AWA2, CUB and SUN under PS data split using two different settings. In the first setting, we follow the existing transductive generalized ZSL (T-GZSL) methods~\cite{paul2019SABR,xian2019f-VAEGAN,wu2020SDGN}, where the testing unseen-class dataset $\mathcal{D}^{U}$ is separately used to train the model and the testing seen-class dataset $\mathcal{D}^{S}$ and testing unseen-class dataset $\mathcal{D}^{U}$ are compounded at the testing stage. In a more strict data setting, the testing seen-class dataset $\mathcal{D}^{S}$ and testing unseen-class dataset $\mathcal{D}^{U}$ are always indistinguishable for a GZSL task. Hence we propose a more strict data setting to evaluate T-GZSL methods. Here we perform GZSL using the first setting for fair comparison with the existing methods. In addition, we present the details in the more strict setting in section 4.5.

In the first setting, since the testing unseen-class dataset $\mathcal{D}^{U}$ could be separately used to train the model, we firstly use the seen-class dataset $\mathcal{D}^{S}_{tr}$ and the testing unseen-class dataset $\mathcal{D}^{U}$ to train an out-of-distribution (OOD) detector~\cite{HendrycksMD19}. We report more details about the OOD detector in the Appendices since it is not our main concern. Then, we train an unseen-class domain classifier using STHS-S2V or STHS-WGAN and train the seen-class domain classifier with seen-class dataset. At the testing stage, the inputs are firstly classified by the OOD detector. Then, the data classified as seen classes and unseen classes are further classified by the seen-class domain classifier and unseen-class domain classifier respectively. We compare the proposed methods with $23$ existing methods. The results are reported in Table~\ref{tab2}. As seen from Table~\ref{tab2}, STHS-S2V and STHS-WGAN both significantly outperform the comparative methods. Specifically, STHS-S2V achieves improvements about $2.7\%$, $1.8\%$, and $1.1\%$ on AWA2, CUB and SUN respectively, while STHS-WGAN achieves improvements about $4.5\%$, $4.9\%$, and $0.2\%$ on AWA2, CUB and SUN respectively. These improvements demonstrate that the proposed STHS framework can select effective unseen-class samples for transductive learning.
\begin{table}[t]
	\centering
	\caption{Comparative results in the generalized ZSL setting on AWA2, CUB and SUN. U = unseen-class ACC, S = seen-class ACC, H = Harmonic mean of unseen-class and seen-class ACCs.}
	\resizebox{0.95\columnwidth}{!}{
		\begin{tabular}{c|c|ccc|ccc|ccc}
			\hline
			& Method& \multicolumn{3}{c}{AWA2}& \multicolumn{3}{c}{CUB}& \multicolumn{3}{c}{SUN} \\
			\cline{3-5} \cline{6-8} \cline{9-11}
			& &  U&  S&  H& U&  S&  H&  U&  S&  H \\
			\hline
			\multirow{12}{*}{$\mathcal{I}$}& f-CLSWGAN&  -& -& -& 43.7& 57.7& 49.7& 42.6& 36.6& 39.4 \\
            & DASCN&       -& -& -& 45.9& 59.0& 51.6& 42.4& 38.5& 40.3 \\
            & DVBE&        63.6& 70.8& 67.0& 53.2& 60.2& 56.5& 45.0& 37.2& 40.7 \\
            & DAZLE&       60.3& 75.7& 67.1& 56.7& 59.6& 58.1& 52.3& 24.3& 33.2 \\
            & OCD-GZSL&    59.5& 73.4& 65.7& 44.8& 59.9& 51.3& 44.8& 42.9& 43.8 \\
            & APNet&       54.8& 83.9& 66.4& 48.1& 55.9& 51.7& 35.4& 40.6& 37.8 \\
            & DE-VAE&      58.8& 78.9& 67.4& 52.5& 56.3& 54.3& 45.9& 36.9& 40.9 \\
            & LsrGAN&     -& -& -& 48.1& 59.1& 53.0& 44.8& 37.7& 40.9 \\
			\hline
            \multirow{11}{*}{$\mathcal{T}$}& ALE-tran&   12.6& 73.0& 21.5& 23.5& 45.1& 30.9& 19.9& 22.6& 21.2 \\
            & GFZSL&     -& -& -& 24.9& 45.8& 32.2& -& -& - \\
            & DSRL&        -& -& -& 17.3& 39.0& 24.0& 17.7& 25.0& 20.7 \\
            & GMN&         -& -& -& 60.2& 70.6& 65.0& 57.1& 40.7& 47.5 \\
            & f-VAEGAN-D2&    84.8& 88.6& 86.7& 61.4& 65.1& 63.2& 60.6& 41.9& 49.6 \\
            & GXE&         80.2& 90.0& 84.8& 57.0& 68.7& 62.3& 45.4& 58.1& 51.0 \\
            & SABR-T&      79.7& 91.0& 85.0& 67.2& 73.7& 70.3& 58.8& 41.5& 48.6 \\
            & PREN&        32.4& 88.6& 47.4& 35.2& 55.8& 43.1& 35.4& 27.2& 30.8 \\
            & VSC&         71.9& 88.2& 79.2& 33.1& 86.1& 47.9& 29.9& 62.9& 40.6 \\
			& SDGN&        88.8& 89.3& 89.1& 69.9& 70.2& 70.1& 62.0& 46.0& 52.8 \\
            & TF-VAEGAN&   87.3& 89.6& 88.4&  69.9& 72.1& 71.0& 62.4& 47.1& 53.7 \\
			\hline
			& STHS-S2V(Ours)&         91.4& 92.3& \textbf{91.8}&  71.2& 74.5& \textbf{72.8}& 70.7& 44.8& \textbf{54.8} \\
			& STHS-WGAN(Ours)&        94.9& 92.3& \textbf{93.6}&  77.4& 74.5& \textbf{75.9}& 67.5& 44.8& \textbf{53.9} \\
			\hline
		\end{tabular}
	}
	\label{tab2}
\end{table}

\begin{figure}[t]
	\centering
	\includegraphics[width=0.9\linewidth]{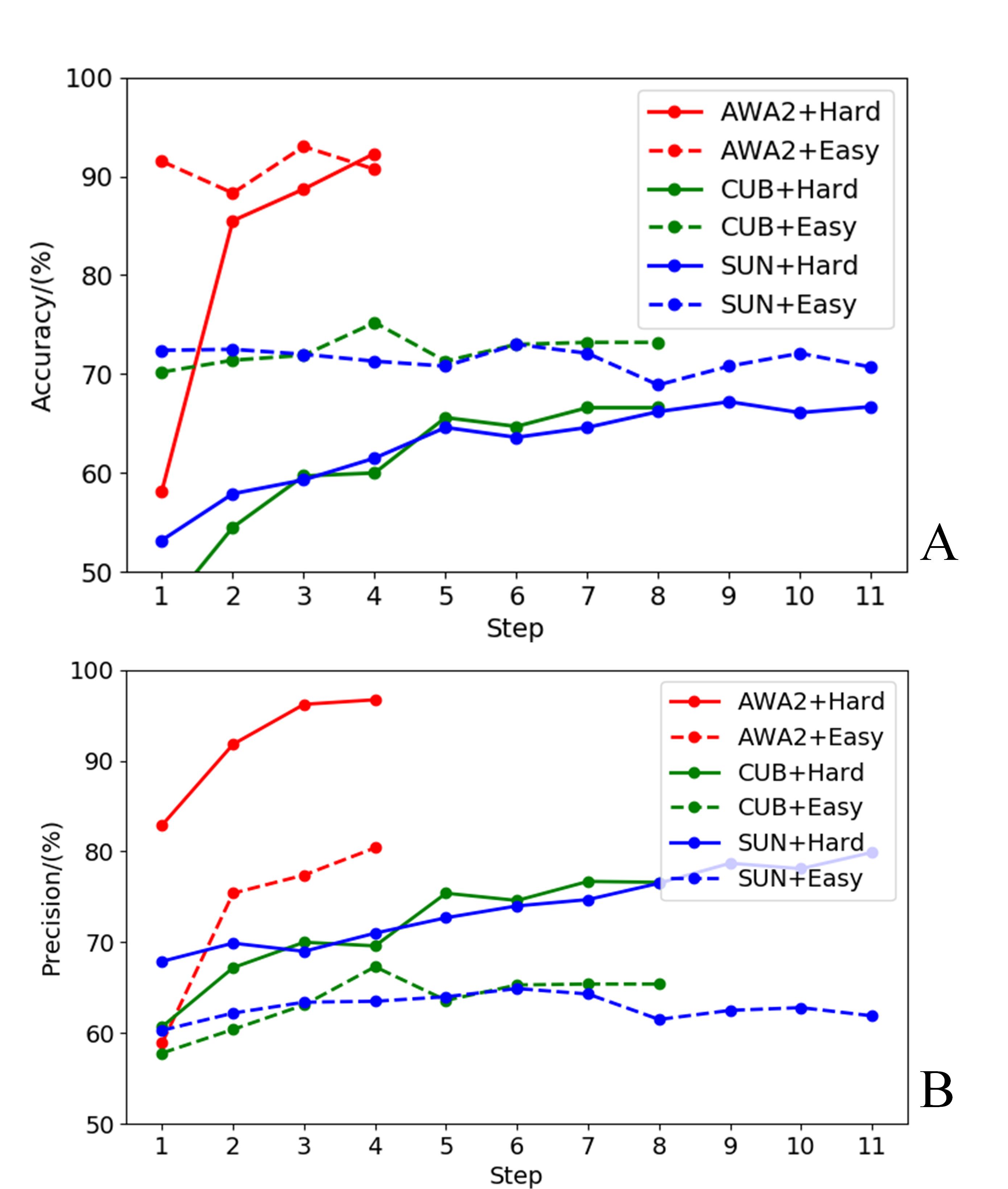}
	\caption{A: the ACCs of hard classes and easy classes in the self-training process. B: the Precisions of hard classes and easy classes in the self-training process.}
	\label{fig3}
\end{figure}

\begin{figure}[t]
	\centering
	\includegraphics[width=0.9\linewidth]{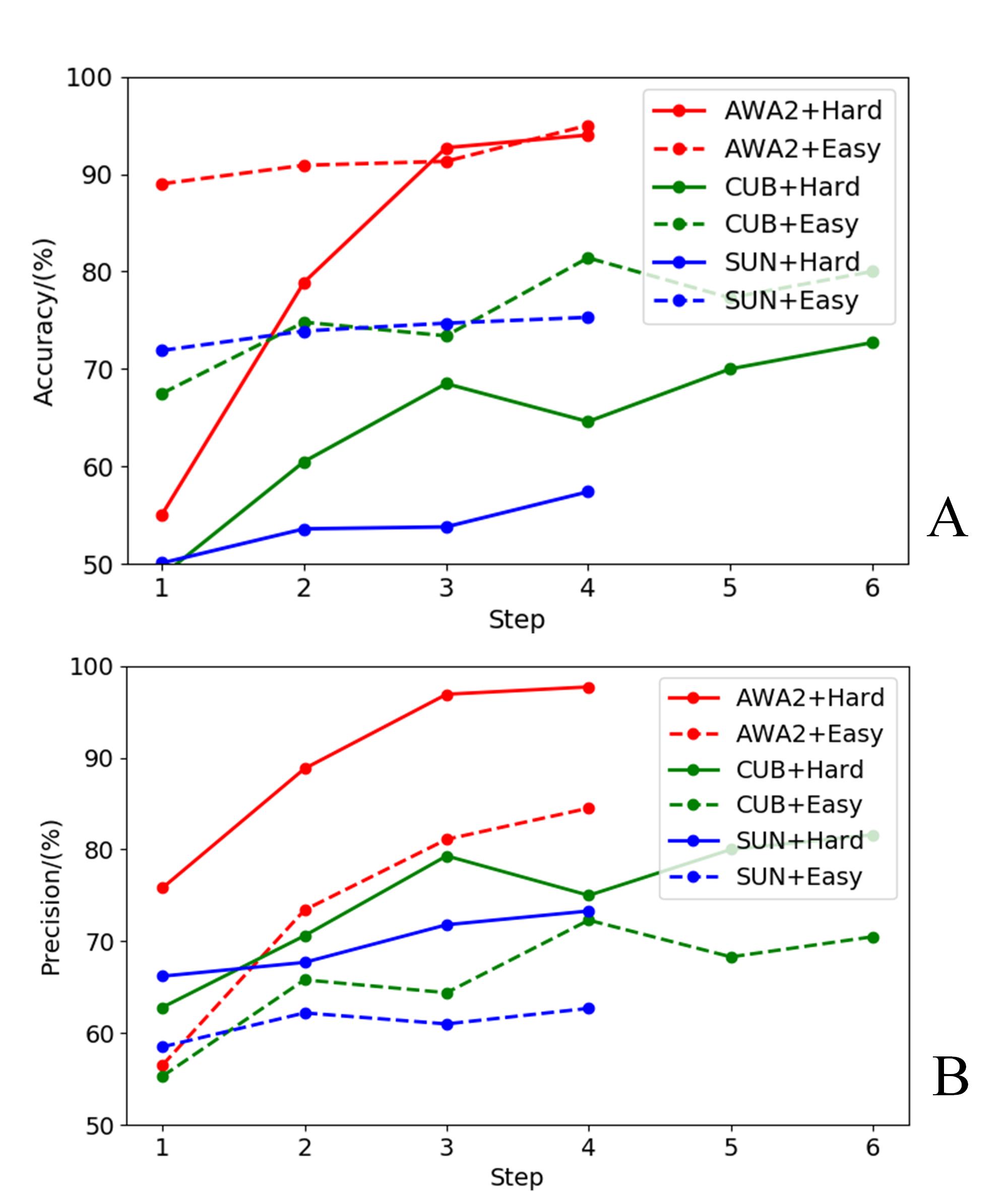}
	\caption{A: the ACCs of hard classes and easy classes in the self-training process. B: the Precisions of hard classes and easy classes in the self-training process.}
	\label{fig4}
\end{figure}

\subsection{Discussions and Ablation}
\textbf{The Identification of Hard Classes.} To show that the proposed two hardness metrics are able to identify the hard classes at each iterative step in the self-training process, we firstly rank the unseen classes from hard to easy via the proposed two hardness metrics (CFBS and PN-CFBS) and compute the per-class Top-$1$ accuracies of unseen classes. We consider the front half of the ranked unseen classes as hard classes and the back half as easy classes. Then we compute the average per-class Top-$1$ accuracies of 1) the hard classes and 2) the easy classes. Specifically, we conduct the experiments using STHS-S2V and STHS-WGAN on AWA2, CUB and SUN in the conventional ZSL setting. AWA2 is used to simulate extremely unbalanced dataset, hence the PN-CFBS metric is used on AWA2 while the CFBS metric is used on both CUB and SUN. The results of STHS-S2V and STHS-WGAN are shown in Figure~\ref{fig3} A and Figure~\ref{fig4} A respectively. From Figure~\ref{fig3} A and Figure~\ref{fig4} A, we can see that the CFBS metric and PN-CFBS metric have the ability to identify the hard classes at each iterative step. In addition, we also find that the difference between accuracies of hard classes and easy classes becomes smaller as the iterative training goes. This indicates that the uneven prediction problem is reduced by the proposed methods.

We also validate our proposed metrics by directly comparing the identified hard classes by our proposed metrics with the `ground-truth' hard classes. Specifically, at each iteration, we could not only identify $50\%$ classes as the hard classes by our proposed metrics, but also determine the $50\%$ `ground-truth' hard classes according to their per-class accuracies. Then, the effectiveness of our proposed metrics is measured by the ratio of the identified hard classes by our proposed metrics to the `ground-truth' hard classes at each iteration. As mentioned above, considering that accuracy gap between hard classes and easy classes gradually narrows as the iterative training goes, the boundary between the hard classes and the easy classes gradually becomes unconspicuous accordingly. Hence we report the results of STHS-S2V at the first three iterations. At the first iteration, this ratio is $4/5$, $18/25$, $25/36$ on AWA2, CUB, SUN respectively; For the second iteration, this ratio is $4/5$, $17/25$, $22/36$ on AWA2, CUB, SUN respectively; And for the third iteration, it is $4/5$, $16/25$, $22/36$ on AWA2, CUB, SUN respectively. These results indicate our metrics are rather effective.

\textbf{The Noise Level of Identified Hard Classes.} To demonstrate that the identified hard classes by the proposed two hardness metrics have less noisy pseudo labels than the easy classes at each iterative step in the self-training process, we compute the average per-class precision (Precision) of 1) the hard classes and 2) the easy classes. The identification of the hard classes and easy classes is accomplished using the method in the above subsection. The experiments are conducted using STHS-S2V and STHS-WGAN on AWA2, CUB and SUN in the conventional ZSL setting. The results of STHS-S2V and STHS-WGAN are shown in Figure~\ref{fig3} B and Figure~\ref{fig4} B respectively. Figure~\ref{fig3} B and Figure~\ref{fig4} B show us that the hard classes identified by the proposed metrics always have significantly less noisy pseudo labels than the easy classes.

\textbf{The Effect of Hardness Sampling.} Here we analyze the effect of the two proposed hardness sampling approaches, CFBS and PN-CFBS. We compute the average per-class Top-$1$ accuracy (ACC) of a model trained with CFBS, ACC of a model trained with PN-CFBS, and ACC of a model trained with random sampling (RS). The experiments are conducted in the conventional ZSL setting on AWA2, CUB and SUN, using STHS-S2V and STHS-WGAN. The results are reported at Table~\ref{tab3}. From Table~\ref{tab3}, we can see that the performances with hardness sampling significantly outperforms those with RS, which demonstrates the effectiveness of the proposed hardness sampling. In addition, Table~\ref{tab3} indicates that PN-CFBS could significantly improve the performance against CFBS when the dataset (AWA2) is extremely unbalanced. We also observe that the performances of PN-CFBS are slightly inferior than those of CFBS on CUB and SUN. On the one hand, this result demonstrates that CFBS has considerable ability to deal with relatively unbalanced dataset. On the other hand, this result indicates that the performance of PN-CFBS is affected by the prior estimation since the prior estimation of a dataset with large number of classes and small number of per-class instances is a hard problem. Anyway, the STHS framework with both CFBS and PN-CFBS could achieve significant improvements against many existing ZSL methods, which demonstrates their effectiveness.

\subsection{A More Strict T-GZSL Data Setting}
In many existing T-GZSL works~\cite{paul2019SABR,xian2019f-VAEGAN,wu2020SDGN}, a popular data setting is that the testing unseen-class dataset $\mathcal{D}^{U}$ is separately used for training, implying that it has been known whether a given sample is an unseen-class(or seen-class) one or not. However, the testing unseen-class dataset $\mathcal{D}^{U}$ and testing seen-class dataset $\mathcal{D}^{S}$ are indistinguishable in a GZSL task. Given this, we propose a more strict data setting to evaluate the T-GZSL methods, where the testing seen-class dataset $\mathcal{D}^{S}$ and testing unseen-class dataset $\mathcal{D}^{U}$ are indistinguishable at both training and testing stages. We generalize the proposed STHS framework to T-GZSL and establish a baseline on this setting. Specifically, we introduce the WGAN model as the base GZSL model and the derived method is denoted as STHS-GZSL. At each iterative step, the model performs GZSL on the compound testing seen-class dataset $\mathcal{D}^{S}$ and testing unseen-class dataset $\mathcal{D}^{U}$ and an unseen-class subset $\mathcal{\bar{D}}^{U}$ is selected with the proposed hardness sampling methods. The experiments are conducted on AWA2, CUB, and SUN. The results are reported at Table~\ref{tab4}. As a contrast, we also report the results of 1) the base I-ZSL model WGAN(Inductive); 2) the WGAN model self-trained with random sampling, named as ST-WGAN-RS(Tansductive). Comparing the results at Table~\ref{tab4} and Table~\ref{tab2}, we can see that the T-GZSL performances decrease significantly when the testing unseen-class dataset could not be separately used to train the model. This decrease is reasonable since the compounded testing seen-class dataset and testing unseen-class dataset will make the transductive training more difficult by, for instance, increasing the noise level of pseudo labels. Even so, the STHS-GZSL still provides a competitive baseline for T-GZSL. In addition, the model using the proposed hardness sampling significantly outperforms the model using random sampling, which demonstrates that the proposed hardness sampling is also effective in T-GZSL.

\section{Conclusion}
In this paper, we propose a Self-Training framework with Hardness Sampling (STHS) for T-ZSL, where the core problem is how to select diverse hard-class samples for iterative training. Addressing this problem, we first empirically analyze the roles of unseen-class samples with different level of hardness in the training process, based on the uneven prediction phenomenon found in many ZSL methods. Then, we propose two hardness sampling techniques which can identify the hard-class samples based on the class frequency and prior-normalized class frequency of model predictions. We introduce two typical I-ZSL methods into the STHS framework and extensive experimental results on three public benchmarks demonstrate that the derived methods significantly outperform many state-of-the-art methods. Besides, we propose a more strict data setting to evaluate T-GZSL and establish a competitive baseline on this setting by generalizing the proposed STHS framework to T-GZSL.
\begin{table}[t]
	\centering
	\caption{Comparative results with different sampling methods on AWA2, CUB and SUN.}
	\resizebox{0.75\columnwidth}{!}{
		\begin{tabular}{c|ccc|ccc}
			\hline
			Method&  \multicolumn{3}{c}{S2V}& \multicolumn{3}{c}{WGAN} \\
			\cline{2-4} \cline{5-7}
				  &  AWA2&  CUB&  SUN&  AWA2&  CUB&  SUN \\
			\hline
			RS&  85.0& 66.0& 66.7& 87.5& 73.1& 63.1 \\
			CFBS&  88.9& \textbf{71.2}& \textbf{70.9}& 89.8& \textbf{77.4}& \textbf{67.5} \\
			PN-CFBS&  \textbf{91.4}& 70.0& 70.0& \textbf{94.9}& 74.5& 65.5 \\
			\hline
		\end{tabular}
	}
	\label{tab3}
\end{table}
\begin{table}[t]
	\centering
	\caption{Comparative results in the more strict T-GZSL data setting on AWA2, CUB and SUN.}
	\resizebox{0.95\columnwidth}{!}{
		\begin{tabular}{c|ccc|ccc|ccc}
			\hline
			Method&  \multicolumn{3}{c}{AWA2}& \multicolumn{3}{c}{CUB}& \multicolumn{3}{c}{SUN} \\
			\cline{2-4} \cline{5-7} \cline{8-10}
				  &  U&  S&  H& U&  S&  H&  U&  S&  H \\
			\hline
			WGAN& 52.2& 76.6& 62.1& 52.9& 46.7& 49.6& 44.0& 30.7& 36.2 \\
			ST-WGAN-RS& 75.1& 82.7& 78.7& 59.7& 68.8& 63.9& 41.9& 35.9& 38.6 \\
			STHS-GZSL& 85.4& 81.0& 83.2& 65.1& 67.4& 66.2& 45.1& 36.2& 40.2 \\
			\hline
		\end{tabular}
	}
	\label{tab4}
\end{table}

\section{Acknowledgements}
This work was supported by the National Natural Science Foundation of China (NSFC) under Grants (61991423, U1805264) and the Strategic Priority Research Program of the Chinese Academy of Sciences (XDB32050100).

{\small
\bibliographystyle{ieee_fullname}
\bibliography{cvpr}
}

\end{document}